# Optical Mapping Near-eye Three-dimensional Display with Correct Focus Cues

WEI CUI,[1,2] AND LIANG GAO[1,2,*]

[1]Department of Electrical and Computer Engineering, University of Illinois at Urbana-Champaign, 306 N. Wright St., Urbana, IL 61801, USA
[2]Beckman Institute for Advanced Science and Technology, University of Illinois at Urbana-Champaign, 405 N. Mathews Ave., Urbana, IL 61801, USA

We present an optical mapping near-eye (OMNI) three-dimensional display method for wearable devices. By dividing a display screen into different sub-panels and optically mapping them to various depths, we create a multiplane volumetric image with correct focus cues for depth perception. The resultant system can drive the eye's accommodation to the distance that is consistent with binocular stereopsis, thereby alleviating the vergence-accommodation conflict, the primary cause for eye fatigue and discomfort. Compared with the previous methods, the OMNI display offers prominent advantages in adaptability, image dynamic range, and refresh rate.

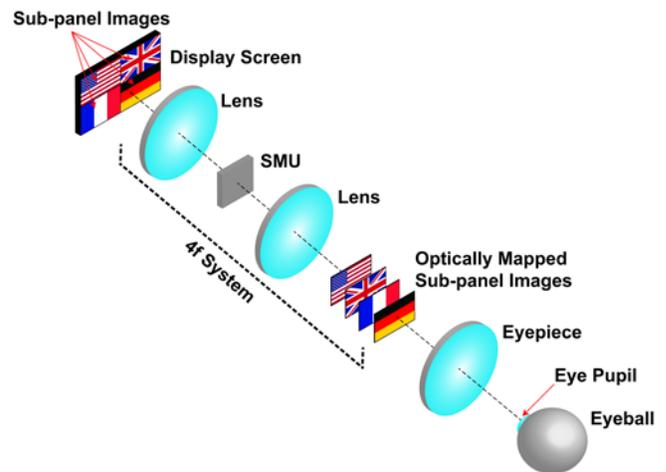

Fig. 1. Operating principle of the optical mapping near-eye (OMNI) three-dimensional display. SMU, spatial multiplexing unit.

Near-eye three-dimensional displays have seen rapid growth and held great promise in a variety of applications, such as gaming, film viewing, and professional scene simulations. Currently, most near-eye three-dimensional displays are based on computer stereoscopy [1], which presents two images with parallax in front of the viewer's eyes. Stimulated by binocular disparity cues, the viewer's brain then creates an impression of the three-dimensional structure of the portrayed scene. However, the stereoscopic displays suffer from a major drawback—the vergence-accommodation conflict [2]—which reduces the viewer's ability to fuse the binocular stimuli while causing discomfort and fatigue. Because the images are displayed on one surface, the focus cues specify the depth of the display screen (*i.e.*, accommodation distance) rather than the depths of the depicted scenes (*i.e.*, vergence distance). This is opposite to the viewer's perception in the real world where these two distances are always the same. To alleviate this problem, one must present correct focus cues that are consistent with binocular stereopsis.

Currently, only a few approaches can provide correct or nearly correct focus cues for the depicted scene, namely light field near-eye displays [3] and multiplane near-eye displays [4]. The light field display employs a lenslet array to project multi-view images simultaneously onto the viewer's retina, thereby yielding a continuous three-dimensional sensation. Despite a compact form factor, the spatial resolution is low (~ 100×100 pixels [5]), restricted by the number of pixels that can fit into the imaging area of a lenslet [6]. By contrast, the multiplane display projects two-dimensional images onto a variety of depth planes through either temporal multiplexing [7] or spatial multiplexing [8]. By synchronizing a fast display with a deformable mirror [3] or a focal sweeping lens [7], the temporal-multiplexing-based methods project depth images in sequence. However, to render continuous motion, the device must display all depth images within the flicker fusion time (1/60 s), thus introducing a severe trade-off between the image dynamic range and the number of depth planes. Alternatively, the spatial-multiplexing-based methods deploy multiple screens at various distances from the viewer, followed by optically combining their images using a beam splitter.

Notwithstanding a high spatial resolution and image dynamic range, the resultant system is generally bulky and therefore unsuitable for wearable applications.

To overcome above limitations, herein we present an optical mapping near-eye (OMNI) three-dimensional display method which provides correct focus cues for depth perception. Based on a similar conceptual thread to the multiplane display, our method maps different portions of a display screen to various depths while forcing their centers aligned. These intermediate depth images are then reimaged by an eyepiece and projected onto the viewer's retina.

The operating principle is shown in Fig. 1. A high-resolution two-dimensional image is displayed at an electronic screen. The image consists of several sub-panels, each targeting to be displayed at a designated depth. The workhorse of our system is a $4f$ optical relay with a spatial multiplexing unit (SMU) located at the Fourier plane. The SMU functions as a multifocal off-axis Fresnel lens, adding both quadratic and linear phase terms to the incident wavefront. The quadratic phase terms axially shift sub-panel images to the designated depths, while the linear phase terms laterally shift the centers of sub-panel images to the optical axis. As a result, the sub-panel images are mapped to different axial locations and laterally aligned at the output end. Finally, the light emanated from these intermediate depth images is collected by an eyepiece and enters the eye pupil. Depending on their relative axial positions, the viewer perceives these multi-depth images at a distance from a near plane to infinity. It is worth noting that, unlike the previous spatial multiplexing approach [2], our system requires only one display screen at the input, thereby maintaining a compact form factor.

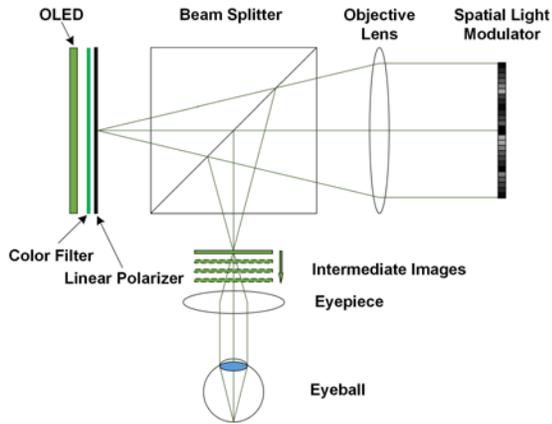

Fig. 2 Optical schematic of an OMNI display using a spatial light modulator. OLED, organic light emitting diode.

Sitting at the heart of an OMNI display, the SMU can be either a reflective or transmissive optical component. In our system, we adopted a reflective configuration and employed a liquid-crystal-on-silicon (LCOS) spatial light modulator (SLM) as the SMU. The optical setup is illustrated in Fig. 2. The light emanated from a monochromatic organic light emitting diode (OLED) screen (MDP02BCYM, 2000×2000 pixels, Microoled) is filtered both in color (central wavelength, 550 nm; bandwidth, 10 nm) and polarization ($p$ light). The filtered light passes through a 50:50 beam splitter, then it is collimated by an infinity-corrected microscope objective (2X M Plan APO, Edmund Optics). We placed the LCOS-SLM at the exit pupil of the objective lens to modulate the phase of the incident light. The reflected light is collected by the same objective lens, reflected at the beam splitter, and forms intermediate images at a variety of depths in front of an eyepiece (focal length, 25 mm; LB1761-A, Thorlabs).

To map the sub-panel image $i$ to the designated location, we need to display a phase pattern on the LCOS-SLM in the form:

$$\varphi_i(x,y) = \frac{\pi(x^2+y^2)}{\lambda f_i} + \frac{2\pi}{\lambda}\left[\sin\left(\frac{l_{x_i}}{f_o}\right)x + \sin\left(\frac{l_{y_i}}{f_o}\right)y\right], \quad (1)$$

where $\lambda$ is the light wavelength, $f_i$ is the effective focal length of the LCOS-SLM, $f_o$ is the focal length of the objective lens in Fig. 2, $l_{x_i}$ and $l_{y_i}$ are center coordinates of sub-panel image $i$ at the OLED. The origin of the coordinate is located at the geometric center of the OLED. The correspondent $f_i$ were calculated and shown in Table 1.

Table 1. Calculated $f_i$ of an OMNI display.

| Dioptric depth (diopter) | 0 | 1 | 2 | 3 |
|---|---|---|---|---|
| $f_i$ (m) | 53.3 | 80.6 | 162.5 | Inf |

Because each sub-panel image requires a different set of $f_i$, $l_{x_i}$, and $l_{y_i}$, the ideal phase pattern that enables simultaneous mapping of all sub-panel images is $\sum_i \varphi_i$. However, in practice, since the displayed phase must be wrapped within $2\pi$ and discretized into 8-bit levels, the simple additive phase pattern is inapplicable. To generate a phase pattern that functions similarly to $\sum_i \varphi_i$, we adopted an optimization algorithm, Weighted Gerchberg-Saxton (WGS) [9, 10]. WGS starts with an initial phase estimate $\varphi_{est}(x,y)$, followed by iteratively updating this estimate to maximize a merit function $T$, which is defined as:

$$T = \sum_{i=1}^{A}\left\{\frac{1}{B}\sum_x\sum_y e^{j[\varphi_{est}(x,y)-\varphi_i(x,y)]}\right\}. \quad (2)$$

Here $x$, $y$ are the discretized Cartesian coordinates, $A$ is the total number of sub-panel images, $B$ is the total number of the LCOS-SLM's pixels, and $j$ is the imaginary number. The optimization process maximizes the overall likelihood between $\varphi_{est}$ and $\varphi_i$ for all sub-panel images.

In addition, to create a three-dimensional scene with continuous depth perception, we employed a linear depth-weighted blending algorithm [11] to create the contents of sub-panel images. In brief, we rendered the image intensity at each depth plane proportional to the dioptric distance of the point from that plane to the viewer along a line of sight. Meanwhile, we maintained the sum of the image intensities a constant at all depth planes.

Table 2. System parameters of an OMNI display.

|  | Lateral resolution (pixels) | Depth plane spacing (diopter) | Frame rate (Hz) |
|---|---|---|---|
| High lateral resolution mode | 1000×1000 | 1.0 | 60 |
| Dense depth sampling mode | 500×500 | 0.2 | 60 |

Compared with existing near-eye three-dimensional displays [3-8], the OMNI display offers prominent advantages in adaptability, image dynamic range, and refresh rate. Because the sub-panel images occupy the same display screen at the input end, the product of a depth plane's lateral resolution ($L\times M$ pixels) and the number of depth planes ($N$) must not be greater than the total number of pixels ($P$) at the display screen, i.e., $L \times M \times N \leq P$. Taking this constraint into consideration, we can configure the system working in two modes which selectively bias the lateral resolution and the depth plane spacing, respectively, simply by alternating the phase patterns on the LCOS-SLM. For the given high-resolution OLED ($P$ = 4 megapixels) and a depth range of 0-3D (diopter), we summarize two typical display settings in Table 2. The scalability of display parameters thus grants us more freedom to adapt the OMNI display to the depicted scene framewise. Furthermore, unlike the temporal-multiplexing-based multiplane display [4], herein the image refresh rate and dynamic range are decoupled from the number of depth planes and thereby limited by only the display screen itself. Using the given OLED, we can display a high dynamic range (8 bits) three-dimensional video in real time (60 Hz).

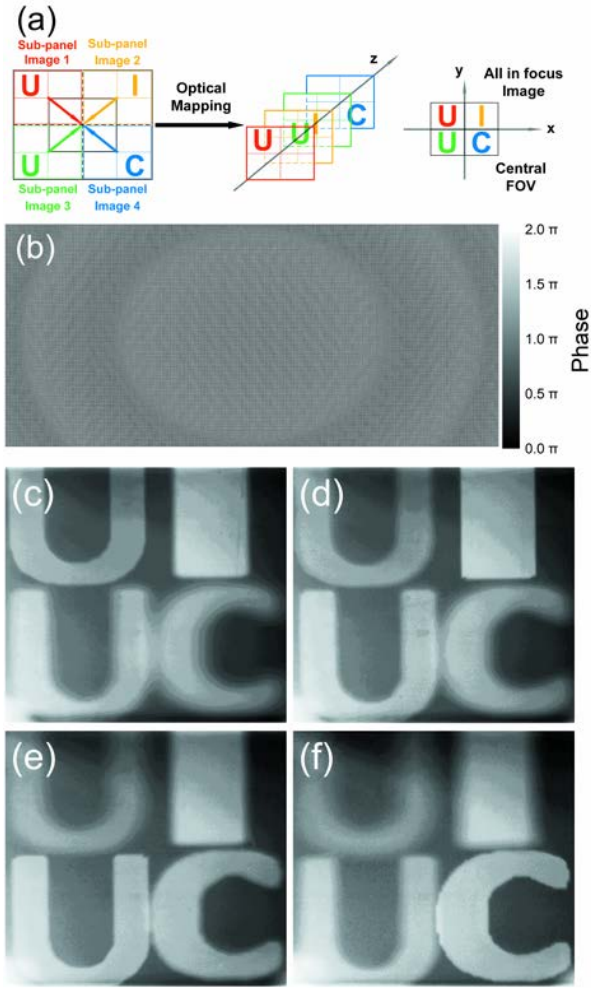

Fig. 3. Visualization of intermediate depth plane images. (a) Optical mapping of four letters "U," "I," "U," "C" from four sub-panels to the central field of view. (b) Optimized phase pattern used in the implementation. (c-f) Intermediate depth images captured at 0D, 1D, 2D, and 3D, respectively.

To visualize the intermediate depth images in the OMNI display, we performed a simple mapping experiment. At the input end, we displayed four letters "U", "I", "U", "C" in the four sub-panels of the OLED (Fig. 3(a)) and set the system working in the high lateral resolution mode (Table 2). The optimized phase pattern (Fig. 3(b)) was employed. We placed a camera at the focus of the eyepiece, translated it towards the eyepiece, and captured images at four nominal depth planes (0D, 1D, 2D, and 3D). The remapped letter images at these four depths are shown in Fig. 3(c)-(f), respectively. As expected, the letters appear sharp at their designated depths while blurred elsewhere.

Next, we varied the depth plane spacing and accommodation distance to evaluate their effects on the image contrast. In Experiment I, we placed a camera in front of the eyepiece to mimic an eye that focused at 1.5D. We displayed two sub-panel images (a slanted edge) at the OLED and projected them to $(1.5+\Delta z/2)$D and $(1.5-\Delta z/2)$D, respectively. We varied the depth plane spacing, $\Delta z$, through changing the effective focal length $f_i$ in the phase pattern at the LCOS-SLM (Eq. 1). Accordingly, we acquired the depth-fused images at the camera, and calculated the image modulation contrast [4]. The dependence of modulation contrast on the depth plane spacing $\Delta z$ is shown in Fig. 4(a). The modulation contrast degrades as the depth plane spacing increases. However, for a given number of depth planes, decreasing the depth plane spacing will unfavorably reduce the total depth range. Therefore, one must balance the image quality for a desired depth range.

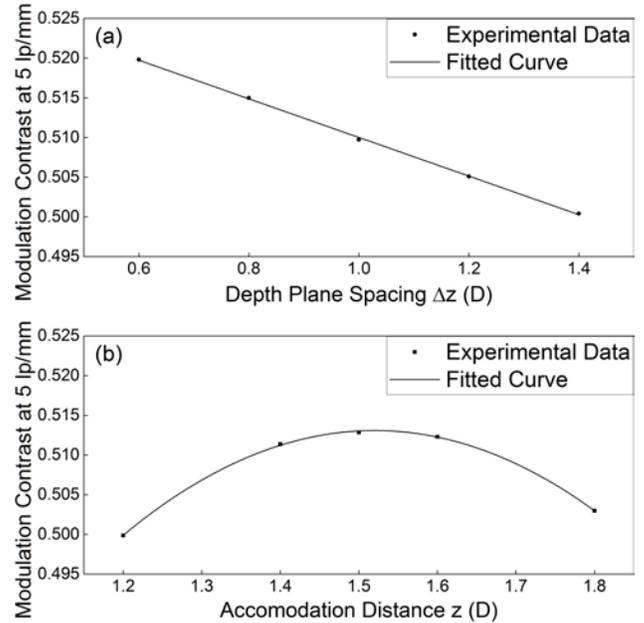

Fig. 4. Evaluation of an OMNI display. (a) Modulation contrast at a given spatial frequency (5 lp/mm) versus depth plane spacing ($\Delta z$). (b) Modulation contrast at a given spatial frequency (5 lp/mm) versus the accommodation distance ($z$). The modulation contrasts were calculated using the standard slanted edge method [4].

In Experiment II, we varied the focal depth, $z$, of the camera to mimic the accommodation distance change of the eye. At the input end, we displayed two identical images at the 1D and 2D depth plane. Because the light intensities along a line of sight at these two depth planes are identical, the rendered depth is at the dioptric midpoint, 1.5D. We captured images at a variety of dioptric accommodation distances and derived the correspondent image modulation contrasts. The result (Fig. 4(b)) shows that the modulation contrast reaches the maximum at $z = 1.5$ D and degrades smoothly around this depth. Although the results shown in Fig. 5 were measured at a representative spatial frequency 5 lp/mm, a similar trend was observed at all other frequencies as well. Since the human eye inherently focuses on the depth that provides the highest modulation contrast, our system thereby provides a correct focus cue that can drive the eye's accommodation to the desired depth. It is noteworthy although the depth spacing (1D) is greater than the minimal depth discrimination of human eye, the perceived depths will still be continuous thanks to the content rendering by the depth blending algorithm.

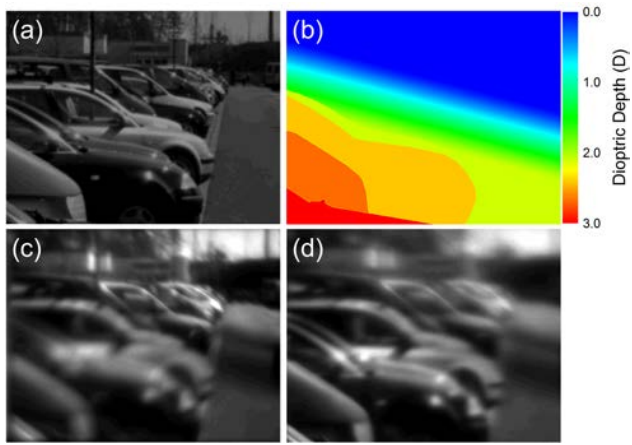

Fig. 6. OMNI display of a complex three-dimensional scene. (a) Ground-truth all-in-focus image. (b) Ground-truth depth map. (c) Representative depth image captured at 0D. (d) Representative depth image captured at 3D.

Last, we tested our system using a complex three-dimensional scene. The ground-truth all-in-focus image and the corresponding depth map are shown in Fig. 6(a) and (b), respectively. We generated the display contents at four nominal depth planes (0D, 1D, 2D, and 3D). Again, we varied the focal depth of the camera to mimic the accommodation distance change of the eye. The depth-fused images captured at a far plane (0D) and a near plane (3D) are shown in Fig. 6(c) and (d), respectively, matching closely with the ground-truth depth map (Fig. 6(b)).

In the OMNI display, we chose an LCOS-SLM as the SMU to accomplish the optical mapping. Although not demonstrated, the SMU can also be other phase modulation devices, such as a volume holography grating [12, 13] or a distorted phase grating [14, 15]. Similar to the LCOS-SLM, both these phase modulators can act as a multifocal off-axis Fresnel lens, directing the sub-panel images to the designated depths while forcing their centers aligned. However, unlike the LCOS-SLM, the volume holograph grating and distorted phase grating are passive devices, a fact that has both pros and cons. On the one hand, passive phase modulators require no power supplies, reducing the system volume as well as power consumption. On the other hand, because their phase patterns are stationary, passive phase modulators cannot scale the display parameters in the adaptive fashion as previously discussed.

The OMNI display can reproduce colors. Using a white light OLED as the input screen, we can further split a sub-panel image into three channels, followed by covering them with a red, green, and blue color filter, respectively. Accordingly, at the LCOS-SLM, we must display a phase pattern that compensates for the wavelength difference, thereby mapping these filtered images to the same depth. Nevertheless, given a desired depth plane spacing, displaying colors will unfavorably reduce the lateral resolution by a factor of three.

In summary, we developed an optical mapping near-eye three-dimensional display method with correct focus cues that are consistent with the binocular vision, thus alleviating the vergence-accommodation conflict. Through mapping different sub-panel images of a display screen to various axial depths, we can create a high-resolution three-dimensional image over a wide depth range. The image dynamic range and refresh rate are limited by only the display screen itself and up to 8 bits and 60 Hz, respectively. In sight of advantages above, we envision the OMNI display method will lead to a new generation of three-dimensional display and exhibit great potentials for various wearable applications.

**Funding.** This work was supported in part by NSF CAREER grant (1652150) and discretionary funds from UIUC. A patent on this prototype is currently pending.

**Acknowledgments.** The authors thank Nuochen Lyu and Minkang Yang for their contributions to the depth-weighted blending and optimization algorithm.

## Full References